\documentclass[10pt,conference]{IEEEtran}

\usepackage{cite}

\usepackage{pgfplots}
\usepackage{booktabs}

\usepackage{pifont}
\newcommand{\cmark}{\ding{51}}%
\ifCLASSINFOpdf
   \usepackage[]{graphicx}
   \graphicspath{{figs/}}
   
   \DeclareGraphicsExtensions{.pdf,.jpeg,.png}
\else
   \usepackage[dvips, pdftex]{graphicx}
   \graphicspath[pdftex]{{../figs/}}
   \DeclareGraphicsExtensions{.eps}
\fi
\usepackage[]{amsmath,amssymb}

\usepackage{amsthm}

\usepackage{algorithmic}

\usepackage{array}

\ifCLASSOPTIONcompsoc
  \usepackage[caption=false,font=normalsize,labelfont=sf,textfont=sf]{subfig}
\else
  \usepackage[caption=false,font=footnotesize]{subfig}
\fi
\usepackage{url}

\makeatletter
\def\ps@IEEEtitlepagestyle{%
  \def\@oddfoot{\mycopyrightnotice}%
  \def\@oddhead{\hbox{}\@IEEEheaderstyle\leftmark\hfil\thepage}\relax
  \def\@evenhead{\@IEEEheaderstyle\thepage\hfil\leftmark\hbox{}}\relax
  \def\@evenfoot{}%
}
\def\mycopyrightnotice{%
  \begin{minipage}{\textwidth}
  \centering \scriptsize
  Copyright~\copyright~2020 IEEE. Personal use of this material is permitted. Permission from IEEE must be obtained for all other uses, in any current or future media, including reprinting/republishing this material for advertising or promotional purposes, creating new collective works, for resale or redistribution to servers or lists, or reuse of any copyrighted component of this work in other works. The original IEEE publication can be accessed at https://ieeexplore.ieee.org/document/9266015
  \end{minipage}
}
\makeatother

\begin{document}
%
\title{A Survey on Deep Learning with Noisy Labels: How to train your model when you cannot trust on the annotations?}

\newif\iffinal
\finaltrue
\newcommand{\cmtid}{99999}


\iffinal




%
\author{\IEEEauthorblockN{Filipe R. Cordeiro\IEEEauthorrefmark{1} and 
Gustavo Carneiro\IEEEauthorrefmark{2}}

\IEEEauthorblockA{\IEEEauthorrefmark{1}Department of Computing, Federal Rural University of Pernambuco, Brazil}
\IEEEauthorblockA{\IEEEauthorrefmark{2}School of Computer Science, Australian Institute for Machine Learning, University of Adelaide, Australia\\Email: filipe.rolim@ufrpe.br, gustavo.carneiro@adelaide.edu.au}
}


\else
  \author{Sibgrapi paper ID: 4 \\ }
\fi

\maketitle

\begin{abstract}
Noisy Labels are commonly present in data sets automatically collected from the internet, mislabeled by non-specialist annotators, or even specialists in a challenging task, such as in the medical field. Although deep learning models have shown significant improvements in different domains, an open issue is their ability to memorize  noisy labels during training, reducing their generalization potential. As deep learning models depend on correctly labeled data sets and label correctness is difficult to guarantee, it is crucial to consider the presence of noisy labels for deep learning training. Several approaches have been proposed in the literature to improve the training of deep learning models in the presence of noisy labels. This paper presents a survey on the main techniques in literature, in which we classify the algorithm in the following groups: robust losses, sample weighting, sample selection, meta-learning, and combined approaches.  We also present the commonly used experimental setup, data sets, and results of the state-of-the-art models. 
\end{abstract}



\IEEEpeerreviewmaketitle

\section{Introduction}

Deep Neural Networks (DNNs) have shown great performance to deal with different computer vision tasks, such as image classification \cite{rawat2017deep}, segmentation \cite{lateef2019survey} and object detection \cite{liu2020deep}, to different areas of applications \cite{brinker2019deep, li2019deep, mahdavifar2019application}. One of the factors that improve the performance of deep learning models is the use of large-scale datasets, such as ImageNet \cite{deng2009imagenet}. Unfortunately, the labeling process of large-scale datasets is expensive and time-consuming, and researchers sometimes resort to cheaper alternatives, such as online queries \cite{xie2019improving} and crowdsourcing \cite{yu2018learning}, which can produce datasets with incorrect or noisy labels. Incorrect labels may also be present in small datasets, where the labeling task is difficult or have divergent opinions between annotators, such as medical images \cite{barkan2019reduce, ma2019blind}.

\begin{figure}[!t]
\centering
\includegraphics[width=0.5\textwidth]{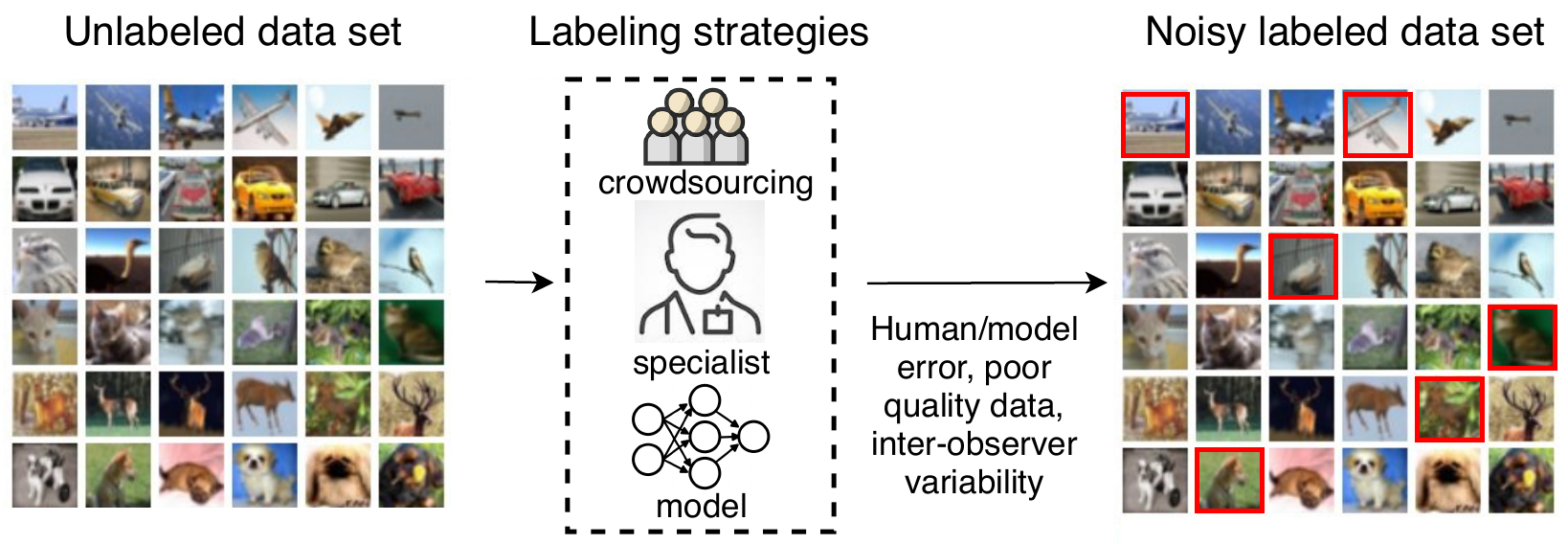}
\caption{Labeling process and noise sources.}
\label{fig:sources}
\end{figure}

As stated in \cite{mcnicol2005primer}, noisy labels may occur naturally when human annotators are involved. Frenay et al. \cite{frenay2013classification} summarize the main sources of noisy labels into four types: 1) insufficient information to provide reliable labeling, such as poor quality images; 2) mistake made by experts; 3) variability in the labeling by several experts; and 4) data encoding or communication problems (e.g., accidental click). Figure \ref{fig:sources} illustrates the labeling strategies and sources of noisy labels.

Most of the solutions using DNN assume that either the labels were annotated by experts or were curated and therefore, would have been perfectly annotated. However, that is not a realistic assumption, mainly when dealing with data collected in an unsupervised way (eg., web queries). As a consequence, a DNN trained with noisy labels might decrease the accuracy and require larger training sets \cite{zhang2018understanding}. 


Noisy labels are also related to semi-supervised learning. As observed by Wang et al. \cite{wang2020proselflc}, when missing labels are incorrectly labeled in a semi-supervised approach, the challenge of semi-supervised learning approximates to noisy labels. The same can be said about pseudo-labeling techniques, where samples can be mislabeled. However, the noisy label problem is even more challenging than previous problems because we do not have information about which samples have clean labels.

Zhang et al. \cite{zhang2018understanding} have shown that Convolutional Neural Networks (CNNs) can easily fit any ratio of noisy labels, leading to poor generalization performance. However, it was shown that easy patterns, corresponding to easy samples with clean labels, are learned first, while difficult patterns, which are closer to noisy labels, are learned later. Based on this observation, many proposed methods are based on the \textit{small-loss trick}, which consists of selecting the samples with small loss to be used as clean samples \cite{jiang2018mentornet, yu2019does, arazo2019unsupervised}. 

Most of the works proposed to deal with noisy labels try to answer the following questions: \textit{How to identify the noisy samples?}, or more generally: \textit{How to effectively train on noisy labeled datasets?} \cite{li2019learning}. To address this problem, different strategies have been proposed: robust losses \cite{wang2019imae, wang2019symmetric}, label cleansing \cite{jaehwan2019photometric, yuan2018iterative}, weighting \cite{ren2018learning}, meta-learning \cite{han2018pumpout}, ensemble learning \cite{miao2015rboost}, and others \cite{yu2018learning, kim2019nlnl, zhang2019metacleaner}. This survey describes the main approaches proposed in the literature related to training deep neural networks in the presence of noisy labels. 



\section{Label noise: Definition and Taxonomy}
\label{sec:definition}

\subsection{Definition}
In this document, we refer to noisy samples as the ones whose  labels are different from their true class. For these samples, we denote their labels as noisy labels, which means their labels are wrong. Therefore, when referring to noisy samples in the scope of this paper, it means that the noise is present only in the labels and not in the input data.

\subsection{Problem statement}

Lets consider a classification problem with a training set $D=\{(x_1, y_1),...,(x_n,y_n)\}$, where $x_i \in \mathcal{X}$ is the $i^{th}$ image and $y_i \in Y$ is a one-hot vector representing the label over $c$ classes. In a noisy label scenario, the labels might be wrong and we denote $y \in Y$ as the observed labels, which may contain noise (i.e., incorrect labels). We denote the true label of $x_i$ as $y_i
^{*}$. We denote the distribution of different labels for sample $x$ by $p(c|x)$, and $\sum_{c=1}^{C} p(c|x) = 1$. 

A supervised classification learns a function $f: \mathcal{X} \to \mathcal{Y}$ that maps the input space to the label space. Training the classifier has as objective to find the optimal parameters $\theta$ that  minimize an empirical risk defined by a loss function. Given a loss function, $L$, and a classifier, $f(.)$, the empirical risk is defined as follows:
\begin{equation}
 R_L(f)=\mathbb{E}_{(x,y)\in D}[L(f(x),y)]=\mathbb{E}_{x,y_x}[L(f(x),y_x)],
\end{equation}
where $\mathbb{E}$ denotes the Monte-Carlo expectation using the training set $D$.

\subsection{Types of noise}

We denote the overall noise rate by $\eta \in [0,1]$ and $\eta_{jc}$ represents the probability of a class $j$ be flipped to class $c$, as $p(y_i=c|y_i^{*}=j)$.
The main types of noise studied in literature are described as follow:

\subsubsection{Symmetric Noise}
Symmetric noise is also called random noise or \textit{uniform} noise, and it represents a noise process when a label has equal probability to flip to another class. Among the symmetric noise definition, there are two variations: \textit{symm-inc} and \textit{symm-exc} \cite{kim2019nlnl}. In \textit{symm-inc}, the true label is included into the label flipping options, which means that in $\eta_{jc}=\frac{\eta}{C-1}, \forall j \in Y$, while in the \textit{symm-exc} the true label is not included, which means $\eta_{jc}=\frac{\eta}{C-1}, j\neq c$. The symmetric noise, or random noise, is unlikely to represent a realistic scenario for noisy labels, but it is the main baseline for noisy labels experiments. Figure \ref{fig:matrix} (a) shows the transition matrix for symmetric noise, with $\eta=0.4$.
\subsubsection{Asymmetric Noise}
\begin{figure}[!t]
\centering
\includegraphics[width=0.4\textwidth]{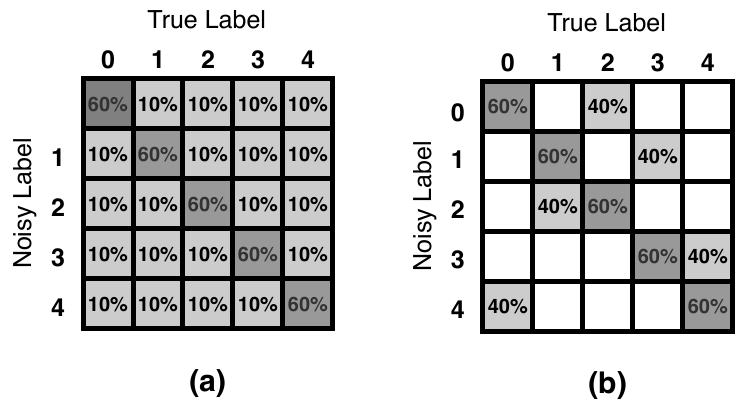}
\caption{Transition matrix of different noisy types: (a) symmetric, and (b) asymmetric, for  $\eta=0.4$ and 5 classes.}
\label{fig:matrix}
\end{figure}

:
The asymmetric noise, as described in \cite{patrini2017making}, is closer to a real-world label noise based on flipping labels between similar classes. For example, using CIFAR10 dataset \cite{krizhevsky2009learning}, the asymmetric noise maps TRUCK$\to$ AUTOMOBILE, BIRD $\to$ PLANE, DEER $\to$ HORSE, as mapped by \cite{zhang2018generalized}. For MNIST, \cite{patrini2017making} maps $2 \to 7$, $3 \to 8$, $7 \to 1$ and $5 \to 6$. For asymmetric noise, $\eta_{jc}$ is class conditional. Figure \ref{fig:matrix} (b) shows the transition matrix for asymmetric noise, with $\eta=0.4$.

\subsubsection{Open-set Noise}: 
The noisy label problem can fall into two categories: closed-set and open-set. A closed-set noisy problem is when all the true labels belong to the known classes. For example, for MNIST, if a subset of samples has the labels flipped to wrong labels, their original images, and corresponding true classes belong to MNIST. 

The open-set noise is when a sample has a true class that is not contained in the known classes of the training data. For example, if an image from CIFAR is contained in MNIST training with an incorrect label of 7, it will be trained as a class 7, but it would be an open-set sample. This type of noise is commonly found when obtaining images from automatic web search engines (e.g., Google Images).

\section{Literature Review}

\label{sec:sota}

Learning in the presence of noisy labels is a problem studied during the last decades \cite{angluin1988learning}, and several strategies have been proposed to make models more robust to noise \cite{frenay2013classification}. In this work, we are grouping the main approaches in the area in the following groups: noise transition matrix, robust losses, sample weighting, sample selection, meta-learning, and combined approaches. Each category is described below:

\subsection{Noise Transition Matrix}

Most of the first approaches proposed to deal with noisy labels were based on estimating a noise transition matrix to learn how labels switch between classes, as illustrated in Figure \ref{fig:matrix}. The cross-entropy loss with transition matrix is defined as follows:
\begin{equation}
\mathcal{L(\theta)} = \frac{1}{N} \sum_{n=1}^{N} -log  ~p(y=y_n|x_n, \theta), 
\end{equation}
where
\begin{equation}
\mathcal{L(\theta)} = \frac{1}{N} \sum_{n=1}^{N} -log (\sum_{i}^{c} p(y=y_n| y^*=i) p(y^*=i|x_n,\theta)).
\end{equation}
Several methods have been proposed to estimate the transition matrix. 
Patrini et al. \cite{patrini2017making} estimate this matrix using a pre-trained model. Hendricks et al. \cite{hendrycks2018using} use a clean validation set to calculate the transition matrix, while Sukhbaatar et al. \cite{sukhbaatar2014learning} propose the use of the difference between the transition matrices calculated from clean and noisy data.




Reed et al. \cite{reed2014training} uses a transition matrix combined with a regularized loss that uses a combination of the noisy labels and labels predicted by the model. Goldberger et al. \cite{goldbergertraining} use the expectation-maximization (EM) algorithm to find the optimal parameters of both network and the noise. The use of transition matrices has been further explored in different ways \cite{chen2015webly, bekker2016training, xia2019anchor}. 

\subsection{Robust Losses}

Loss correction approaches usually add a regularization or modify the network probabilities to penalize less the low confident predictions, which may be related to noisy samples. One of the advantages of these approaches is that they can be used with any model. Most of the methods treat the noisy and clean samples the same way, but penalise less the low confident prediction samples, compared to the standard cross-entropy \cite{liu2020deep}. 

Manwani et al. \cite{manwani2013noise} show that  0-1 losses are more noise-tolerant than commonly used convex losses. \cite{ghosh2017robust} compare categorical cross-entropy (CCE) with mean absolute value of error (MAE) losses, and show that MAE is more noise tolerant because MAE treats all data points equally. However, training with MAE usually leads to underfit, and it may not be beneficial using it depending on the noise type. Zhang and Sabuncu \cite{zhang2018generalized} propose a generalized cross-entropy loss by combining the benefits of mean absolute error and cross-entropy losses. Wang et al. \cite{wang2019imae} propose an improved version of MAE (IMAE), which uses hyperparameters to adjust the weighting variance of MAE. 
Wang et al. \cite{wang2019symmetric} propose the symmetric cross-entropy, based on the fact that CCE is not symmetric. By adding symmetry to the loss, they show that it helps to deal with noisy labels. 
Ziyin et al. \cite{ziyin2020learning}  propose the use of a loss function that encourages the model to abstain from learning samples with noisy labels. This idea is similar to re-weighting or sample selection, where the noisy samples can be observed as having zero weight or removed. Similarly, \cite{thulasidasan2019combating} also proposes a loss function that permits abstention during training. Although some approaches benefit from removing noisy samples, using them through relabelling or giving a lower weight has shown to improve results.

Ma et al. \cite{ma2020normalized} propose a robust loss function called Active Passive Loss (APL) that combines two robust loss functions that mutually boost each other. Ma et al. identify that existing robust loss functions can deal with noisy labels, but suffer from the problem of underfitting. Their proposal addresses this problem by combining losses that cause overfit, such as CCE, with one that causes underfit, as MAE.

Liu et al. \cite{liu2019peer} propose a peer loss function inspired in a peer prediction mechanism. They show that peer loss functions on the noisy data lead to the optimal or a near-optimal classifier as if performing training over the clean training data.

\subsection{Sample Weighting}

Wang et al. \cite{wang2018iterative} propose a weighting scheme to reduce the contribution of the noisy samples. Similarly, \cite{guo2018curriculumnet} uses a method based on Curriculum Learning, which defines a weight to each sample based on an unsupervised estimation of data complexity.

Xue et al. \cite{xue2019robust} use a probabilistic Local Outlier Factor algorithm (pLOF), which is used as an outlier detector,  to estimate a probability value of a sample be an outlier (i.e., noisy sample). \cite{wang2018iterative} also uses the pLOF, but combined with a Siamese Network training. Using siamese networks encourages the model to learn similar features between clean samples of the same class and different ones among clean and noisy samples. 

Harutyunyan et al. \cite{harutyunyan2020improving} show that the memorization of label noise can be reduced by reducing the mutual information between weights. In their proposal, they update the weights based on the gradients of final layers, without accessing the labels. 

Lee et al. propose the CleanNet~\cite{lee2018cleannet}, which uses a predefined subset of reference images. The visual features of the reference subset are extracted using autoencoder and each new sample for training is compared with the features from the reference set. Based on the distance, a weight is set for each sample and the weighted cross-entropy is calculated. 

\subsection{Sample Selection}

Jiang et al. \cite{jiang2018mentornet} proposes MentorNet, which uses a curriculum scheme by learning first the samples, which are probably correct. MentorNet learns a data-driven curriculum dynamically with a second network, called StudentNet.

Co-teaching is proposed by Han et al. \cite{han2018co} and trains two deep models simultaneously, and each network selects the batch of data for each other, based on the samples with a small loss. Later, they propose Co-teaching+ \cite{yu2019does}, that uses the samples with a small loss, but that disagree on the predictions, to select the data to each other. Wei et al. \cite{wei2020combating} uses the same idea of CoTeaching, but it uses the joint agreements instead of disagreement and calculates a joint loss with Co-Regularization.

Nguyen et al. propose an algorithm called SELF \cite{nguyen2020self}. SELF uses a filtering mechanism based on a model ensemble learning the remove the noisy samples from the supervised training. However, they still use the removed samples to leverage the learning using an unsupervised loss. The ensemble mechanism is based on the model's predictions in different epochs, using an exponential moving average of model snapshots. Different from most of the state of the art methods, SELF requires a small clean validation set to perform well.

\subsection{Meta Learning}

The methods described here use as main approach Meta-Learning models to deal with noisy labels. Although in the end they use meta-learning to reweight or filter samples, we grouped them here because they use a different approach to address the problem.

Ren et al. \cite{ren2018learning} use a Meta-Learning paradigm to reweight the training samples based on their gradients directions. They perform a meta gradient descent step to minimize the loss on a clean validation set. In \cite{shu2019meta} it is also used a meta-learning approach with a clean validation set, but they use a multi-layer perceptron to learn a loss-weighting function.

Li et al. \cite{li2019learning} propose optimize a meta-objective before conventional  training. By generating synthetic noisy labels, they aim to optimize a model that does not overfit to a wide
spectrum of artificially generated label noise. By training the model in several generated synthetic noisy, they aim to have a noise-tolerant model able to consistently learn the underlying knowledge from data despite different label
noise.

\subsection{Combined Approaches}

Mixup \cite{zhang2017mixup} is a technique proposed for data augmentation, that uses a linear combination between samples and labels. Their paper shows that their method is noise-tolerant, and recent literature methods started to incorporate Mixup as an important part of their algorithms.
Zhang et al. \cite{zhang2020distilling} combine the reweighting approach using meta-learning, proposed by \cite{ren2018learning}, with pseudo-label estimation and Mixup. Although they achieve state of the art for high noise regimes, it requires a small clean validation set. 

Kim et al. \cite{kim2019nlnl} propose the use of the learning method called Negative Learning (NL), where it is used as a complementary label (i.e., a label that is different from the annotation). By using complementary labels, the chances of selecting a true label as a complementary are low, and NL decreases the risk of providing incorrect information. Therefore, training with NL is more robust to noise. However, this approach requires a longer training time, and the chance of providing incorrect information increases as the number of classes in the problem increases. In their approach, they propose three stages: Negative Learning, Positive Learning (standard training), and fine-tuning with relabeled samples.

Han et al.~\cite{han2019deep} propose to estimate class prototypes using an iterative training divided into two stages. The first stage train the network with the original noisy labels and the modified labels from the second stage. The second stage uses the trained network from the first stage and refines the prototypes, relabelling samples. With the automatic identification of prototypes, it does not need the use of a clean auxiliary set.  

The DivideMix approach \cite{li2020dividemix} combines several methods used in literature to deal with the noisy label problem. It first separates the samples in clean and noise based on Arazo's approach \cite{arazo2019unsupervised}, using Mixture of Gaussians. At the same time, it uses the co-training strategy, where it trains two networks at the same time to avoid error accumulation. After it splits the data in clean and noisy, it trains the model in a semi-supervised approach, using MixMatch \cite{berthelot2019mixmatch}. The whole process is repeated at each epoch.

The main approaches described in this section are summarized in Table \ref{tab: approaches} that shows the combined strategies used by the methods of the state-of-the-art. Table \ref{tab:comb} shows the main components of the combined techniques in the state-of-the-art and the difference between them.


\begin{table*}[ht]
\centering
\caption{Main approaches in the literature to deal with noisy labels.}
\label{tab: approaches}
\begin{tabular}{ccp{5cm}p{5cm}}

\toprule
Approaches & Methods & Advantages & Disadvantages \\
\midrule
Transition Matrix & \cite{patrini2017making, hendrycks2018using, sukhbaatar2014learning, reed2014training, goldbergertraining, chen2015webly, bekker2016training, xia2019anchor} & Easy to implement. & Difficult and complex to estimate the transition matrix in practice. \\
Robust Losses & \cite{manwani2013noise, ghosh2017robust, zhang2018generalized, wang2019imae, wang2019symmetric, ziyin2020learning, thulasidasan2019combating, ma2020normalized, liu2019peer} & It can be easily added to any training model. & Requires to be combined with other strategies to be competitive with state-of-the-art.\\
Sample Weighting & \cite{guo2018curriculumnet, xue2019robust, wang2018iterative, harutyunyan2020improving, lee2018cleannet} & Reduces the influence of noisy samples, but still uses information from it. & Hard to define a correct weighting without the need of a clean auxiliary set.\\
Sample Selection & \cite{jiang2018mentornet, han2018co, yu2019does, wei2020combating, nguyen2020self} & Filter clean samples.  &  Not competitive with state-of-the-art because do not use the noisy samples in an unsupervised way.\\
Meta-Learning & \cite{ren2018learning, li2019learning, shu2019meta} & Has big potential of generalization among different tasks. & Usually requires a clean validation set.\\
Combined & \cite{zhang2017mixup, zhang2020distilling, ren2018learning, kim2019nlnl, li2020dividemix, arazo2019unsupervised} & Good performance, being the state-of-the-art. & Use a set of combined methods that adds complexity to the solution. \\

\bottomrule 
\end{tabular} 
\end{table*}

\begin{table*}[]
\caption{State-of-the-art approaches and combined techniques}
\label{tab:comb}
\centering
\begin{tabular}{ccccccccc}
\toprule
Method & Mixup (Data aug.) & Weighting & filtering & robust loss & regularization& Emsemble & Pseudo-labeling & val. set\\
\midrule
 NLNL \cite{kim2019nlnl}       &   &   &   & \cmark  & \cmark  &   & \cmark  &  \\
 SELF \cite{nguyen2020self}       &   &   & \cmark  &\cmark   &   &   & \cmark & \cmark   \\
 M-correction \cite{arazo2019unsupervised}      &   &   & \cmark  &   & \cmark  &   &   &\\
 Zhang \cite{zhang2020distilling} & \cmark  & \cmark  &   &   & \cmark  &   &  &  \cmark\\
Coteaching+ \cite{yu2019does} &   &   & \cmark  &   &   & \cmark  &  &    \\
 DivideMix \cite{li2020dividemix}  & \cmark  &   & \cmark  &   & \cmark  &  \cmark &   &   \\
 \bottomrule
\end{tabular}
\end{table*}

\section{Experimental Setup}
\label{sec:method}

Most of the papers in the literature evaluate their methods by generating synthetic noise on commonly used data sets and by using data set with  images collected from the internet, which may contain closed-set and open-set noise. The most used data sets used for robust model evaluations in literature are CIFAR-10/CIFAR-100~\cite{krizhevsky2009learning}, Clothing1M~\cite{xiao2015learning}, Webvision~\cite{li2017webvision} and Food101-N~\cite{lee2018cleannet}.

CIFAR-10 has 10 classes with 5000 32$\times$32 pixel training images per class (forming a total of 50000 training images), and a testing set with 10000 32$\times$32 pixel images with 1000 images per class. CIFAR-100 has 50000 training images, but with 100 classes with 5000 32$\times$32 pixel images per class. This data set was curated and it is assumed not to have any noisy label. Therefore, it is a common approach to add synthetic noise and evaluate the robustness of the model in a noisy data set compared to the original clean version.

Clothing1M consists of 1 million training images acquired from online shopping websites, with labels generated by surrounding texts provided by sellers.  The images from the data set may vary in size, but a common approach is to resize the images to 256$\times$256 for training. This data set contains real-world error on the labels and it is composed of 14 classes.

The Webvision contains 2.4 million images collected from the internet, with the same classes from ILSVRC12, from ImageNet~\cite{deng2009imagenet}. The images from the data set are not all with the same size, being a common approach to resize the images to 256$\times$256 for training. Although ImageNet has 1000 classes, most of the papers use only the 50 first classes for training, because they contain most part of the the noise. The evaluation from the models trained using Webvision is usually done using both Webvision and ILSVRC12 validation sets.

Food101-N~\cite{lee2018cleannet}  is an image data set containing about 310009 training images of food recipes classified in 101 classes and 25000 images for the testing set. As the image size vary may vary, a common approach to resize the images to 256$\times$256 for training. Its also a real-world data set, with an estimated noise rate of 20\%, and it is based on the Food101 data set \cite{bossard2014food}, but it has more images and it is more noisy. 

Table \ref{datasets} shows the main information about the most used data sets for evaluation of solutions in noisy label environments.

\begin{table}[htp]
\scriptsize
\label{datasets}
\caption{Main data sets used in literature for noisy labels.}
\begin{tabular}{ccccc}
\toprule
Data set& \# of training & \# of testing  & \# of class &   estimated noise\\
\midrule
CIFAR-10~\cite{krizhevsky2009learning} & 50000 & 10000 & 10 &  0\%\\
CIFAR-100~\cite{krizhevsky2009learning} & 50000 & 10000 & 100 &  0\%  \\
Clothing1M~\cite{xiao2015learning} & 1M & 10000 & 14 &  38.46\% \\
Webvision~\cite{li2017webvision} & 1M & 50000 & 1000  & 20\% \\
Food101-N~\cite{lee2018cleannet} & 310000 & 55000 & 101  &19.66\%\\
\bottomrule
\end{tabular}
\end{table}


\section{State-of-the-art Results}
\label{sec:results}

\begin{figure}[!ht]
\centering
\includegraphics[width=2.0in]{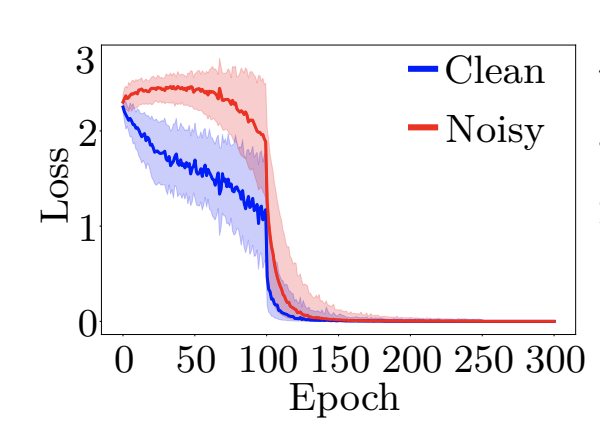}
\caption{Cross-Entropy loss of clean and noisy samples, for 80\% noisy rate, for CIFAR-10. Figure from Arazo's paper \cite{arazo2019unsupervised}.}
\label{fig:arazo}
\end{figure}

Most of the state-of-the-art methods use the fact that CNN tends to learn easy patterns first and then fit the hardest ones \cite{Zhang2017UnderstandingDL}. Arazo et al. \cite{arazo2019unsupervised} showed in his paper how the values of loss are different among clean and noisy samples. Figure \ref{fig:arazo} shows the behavior of loss function for clean and noisy, reported in \cite{arazo2019unsupervised}. The strategy of filtering the samples based on loss is called \textit{small trick} and has been exploited to identify clean and noisy samples. However, the main problem is that hard samples with correct labels can behave like noisy samples, and at the same time, some noisy samples can behave like a clean sample. Therefore, this approach can not be used to filter all samples, but it helps identify most of the noise. Arazo et al. proposes the use of a Gaussian Mixture Model (GMM) to separate the clean and noise during the training, based on the loss value of each sample. For the samples predicted as clean, it uses standard training, with regular cross-entropy, while the the samples predict as noisy are used to regularize the model in an unsupervised way.

\begin{figure}[!t]
\centering
\subfloat[Epoch 1]{\includegraphics[width=1.7in]{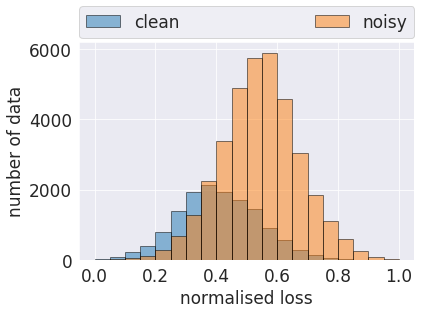}%
\label{fig_first_case}}
\hfil
\subfloat[Epoch 300]{\includegraphics[width=1.75in]{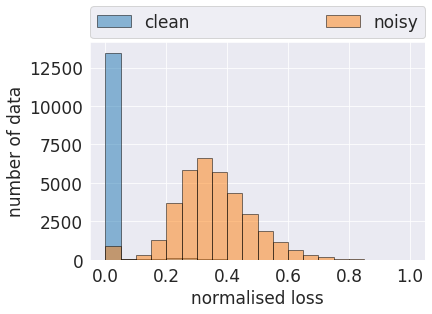}%
\label{fig:hist}}
\caption{Loss histogram for clean and noisy samples, using DivideMix, for 80\% noise rate, symmetric noise,  for CIFAR-10.  }
\label{fig_sim2}
\end{figure}

DivideMix achieves a better split of clean and noisy samples by using the \textit{small trick} combined with Mixup algorithm. Figure \ref{fig_sim2} shows the results of the split for 80\% symmetric noise rate for CIFAR-10. 

DivideMix is currently the state-of-the-art for noisy labels, considering symmetric and asymmetric closed-set noise, without requiring a clean validation set. As described in section III, DivideMix is a combination of methods, which combines Arazo's approach, using GMM to separate clean and noisy samples, Mixup data augmentation and co-training strategy. It also uses standard data augmentation, such as rotation and flipping, as most of the methods described. The main idea of DivideMix is to separate the clean and noisy samples, using GMM, and then treat the problem as a semi-supervised problem, using MixMatch algorithm, that is a variation of the Mixup method proposed for semi-supervised problem. Furthermore, the co-training strategy helps with the noisy training.

As different methods in literature use different model architectures and sometimes different data sets, it is hard to make a fair comparison between most of them. In Table \ref{tab:dmix}  we show the state-of-the-art for CIFAR-10, CIFAR-100, Webvision and Clothing1M, using PreActResNet18 (PRN18). We did not include the methods which use an auxiliar clean validation set, such as in \cite{nguyen2020self} and \cite{zhang2020distilling}, to make a fair comparison, and only the methods that use PRN18 in the original paper. Table \ref{tab:sota_cl} shows the SOTA results for Clothing1M. Table \ref{tab:sota_web} shows the SOTA results for Webvision and ImageNet. Table \ref{tab:food} shows the SOTA results for Food-101N. All the results are the ones reported in the original papers. The original code for reproduction of the results are also available and reported in original papers.

\begin{table}[ht]
\scriptsize
\caption{SOTA results for CIFAR-10 and CIFAR-100, using PRN18. Comparison results adapted from \cite{li2020dividemix}}. 
\label{tab:dmix}
\begin{tabular}{p{2cm}|p{0.2cm}p{0.2cm}p{0.2cm}p{0.3cm}|p{0.4cm}||p{0.2cm}p{0.2cm}p{0.2cm}p{0.2cm}}
\toprule
Data set & \multicolumn{5}{c}{CIFAR-10} & \multicolumn{4}{c}{CIFAR-100}\\    
\midrule
Noise type & \multicolumn{4}{c}{sym.} & asym. &  \multicolumn{4}{c}{sym.} \\
\midrule

Method/ noise ratio & 20\% & 50\% & 80\% & 90\% & 40\%& 20\% & 50\% & 80\% & 90\% \\
\midrule
Cross-Entropy \cite{li2020dividemix} & 86.8 & 79.4 & 62.9 & 42.7 & 85.0 & 62.0 & 46.7 & 19.9 & 10.1\\
Coteaching+ \cite{yu2019does} & 89.5 & 85.7 & 67.4& 47.9& - &65.6& 51.8 & 27.9 & 13.7\\
Mixup \cite{zhang2017mixup} & 95.6 & 87.1 & 71.6& 52.2& - & 67.8& 57.3 & 30.8 & 14.6\\
PENCIL \cite{yi2019probabilistic}& 92.4 & 89.1 & 77.5& 58.9& 88.5 & 69.4& 57.5 & 31.1 & 15.3 \\
Meta-Learning \cite{li2019learning}& 92.9 & 89.3 & 77.4& 58.7& 89.2 & 68.5& 59.2 & 42.4 & 19.5 \\
M-correction \cite{arazo2019unsupervised}& 94.0 & 92.0 & 86.8& 69.1& 87.4 & 73.9& 66.1 & 48.2 & 24.3 \\
DivideMix \cite{li2020dividemix}& \textbf{96.1} & \textbf{94.6} & \textbf{93.2} & \textbf{76.0}& \textbf{93.4} & \textbf{77.3} & \textbf{74.6} & \textbf{60.2} & \textbf{31.5} \\
\bottomrule
\end{tabular}
\end{table}

\begin{table}[ht]
\caption{SOTA results for Webvision and ImageNet ILSVRC12. Results are adapted from \cite{li2020dividemix}.}
\centering
\label{tab:sota_web}
\begin{tabular}{llll}
\toprule
Method & Webvision & ILSVRC12  \\
\midrule
 F-correction~\cite{patrini2017making}    & 61.12 & 57.36  \\
 MentorNet~\cite{jiang2018mentornet}      & 63.00 & 57.80  \\
 Co-teaching~\cite{han2018co}    & 63.58 & 61.48 \\
 Iterative-CV~\cite{chen2019understanding}   & 65.24 & 61.60 \\
 DivideMix~\cite{li2020dividemix}      & \textbf{77.32} & \textbf{75.20}\\
 \bottomrule
\end{tabular}
\end{table}

\begin{table}[ht]
\caption{SOTA results for Clothing1M. Results are adapted from \cite{li2020dividemix}.}
\centering
\label{tab:sota_cl}
\begin{tabular}{ll}
\toprule
Method & Test Accuracy \\
\midrule
 Cross-Entropy~\cite{li2020dividemix}  & 69.21  \\
 M-correction \cite{arazo2019unsupervised}   & 71.00 \\
 PENCIL\cite{yi2019probabilistic} &   73.49 \\
 DeepSelf~\cite{han2019deep} & 74.45 \\
 CleanNet~\cite{lee2018cleannet}    & 74.69 \\
 DivideMix~\cite{li2020dividemix}      & \textbf{74.76} \\
 \bottomrule
\end{tabular}
\end{table}

\begin{table}[ht]

\caption{SOTA results for Food-101N.}
\centering
\label{tab:food}
\begin{tabular}{ll}
\toprule
Method & Food101-N \\
\midrule
 Cross-Entropy\cite{lee2018cleannet}  & 81.44  \\
 CleanNet \cite{lee2018cleannet}   & 83.95 \\
 DeepSelf \cite{han2019deep} & \textbf{85.11} \\

 \bottomrule
\end{tabular}
\end{table}

\section{Conclusion}
\label{sec:conclusion}

Several studies have been proposed in the literature to address the noise label problem. Different strategies have been investigated to make the training of deep learning models more robust to noise labels. Combined strategies based on data augmentation, robust loss, sample filtering, and semi-supervised approaches are currently state-of-the-art. Although we have seen an increasing interest in noisy label problems, there is still much room for improvement, mainly related to asymmetric noise and open-set noise.

Addressing the noisy label problem also impacts other areas, such as pseudo-labeling, semi-supervised and unsupervised training, where recent proposals use predicted labels to improve the training, and these labels can potentially be incorrect. At the same time, many recent strategies from these fields are also applied to noisy labels.

Recent advances have shown that the loss values of samples, mainly at the beginning of training, can help separate the clean and noisy samples. Moreover, data augmentation strategies, such as Mixup, can prevent the model from easily memorizing the noisy samples. However, it is an open question of how to differentiate hard clean samples from noisy samples. Also, semantic noise and open-set noise must be more investigated in future works.






\bibliographystyle{IEEEtran}
\bibliography{main}
%
%


\end{document}